%% file: rnn-hls4ml.tex
\documentclass{article}

\usepackage{arxiv}

\usepackage[utf8]{inputenc} 
\usepackage[T1]{fontenc}    
\usepackage{hyperref}       
\usepackage{url}            
\usepackage{booktabs}       
\usepackage{amsfonts}       
\usepackage{amsmath}
\usepackage{nicefrac}       
\usepackage{microtype}      
\usepackage{lipsum}
\usepackage{graphicx}
\usepackage{xcolor}
\usepackage{subcaption}
\usepackage{verbatimbox}
\usepackage{float}
\usepackage{multirow}
\usepackage[sort,compress,numbers]{natbib}
\usepackage{lineno}
\usepackage{xspace}
\graphicspath{ {./} }

\newcommand{\kt}{\ensuremath{k_{\mathrm{T}}}\xspace}
\newcommand{\pt}{\ensuremath{p_{\mathrm{T}}}\xspace}
\newcommand{\GeV}{\ensuremath{\,\text{Ge\hspace{-.08em}V}}\xspace}
\newcommand{\TeV}{\ensuremath{\,\text{Te\hspace{-.08em}V}}\xspace}

\title{Ultra-low latency recurrent neural network inference on FPGAs for physics applications with hls4ml}

\author{
Elham E Khoda\textsuperscript{1}\thanks{ekhoda@uw.edu}~,
Dylan Rankin\textsuperscript{2}\thanks{drankin@mit.edu}~,
Rafael Teixeira de Lima\textsuperscript{3}\thanks{rafaeltl@slac.stanford.edu}~ \\
\textbf{Philip Harris\textsuperscript{2}, 
Scott Hauck\textsuperscript{1}, 
Shih-Chieh Hsu\textsuperscript{1}, 
Michael Kagan\textsuperscript{3}, 
Vladimir Loncar\textsuperscript{4}}, \\
\textbf{Chaitanya Paikara\textsuperscript{1},
Richa Rao\textsuperscript{1},
Sioni Summers\textsuperscript{4},
Caterina Vernieri\textsuperscript{3},
Aaron Wang\textsuperscript{1}}\\~\\
$^{1}$University of Washington,
Seattle, WA 98195, USA\\
$^{2}$Massachusetts Institute of Technology
  Cambridge, MA 02139, USA \\
$^{3}$SLAC National Accelerator Laboratory,
Menlo Park, CA 94025, USA\\
$^{4}$European Organization for Nuclear Research (CERN), Geneva, Switzerland
}

\date{}

\begin{document}
\begin{center}
\includegraphics[width=8cm]{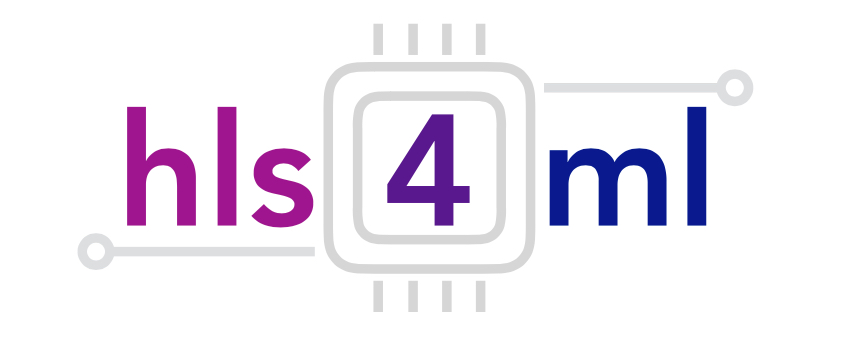}
\end{center}
\maketitle

\begin{abstract}
Recurrent neural networks have been shown to be effective architectures for many tasks in high energy physics, and thus have been widely adopted. 
Their use in low-latency environments has, however, been limited as a result of the difficulties of implementing recurrent architectures on field-programmable gate arrays (FPGAs). 
In this paper we present an implementation of two types of recurrent neural network layers---long short-term memory and gated recurrent unit---within the hls4ml framework.
We demonstrate that our implementation is capable of producing effective designs for both small and large models, and can be customized to meet specific design requirements for inference latencies and FPGA resources.
We show the performance and synthesized designs for multiple neural networks, many of which are trained specifically for jet identification tasks at the CERN Large Hadron Collider.
\end{abstract}

\input{text/01_intro}
\input{text/02_relatedwork}

\input{text/03_impl}

\input{text/04_benchmarks}

\input{text/05_top}


\input{text/06_ftag}

\input{text/07_qdraw}


\input{text/08_results}

\input{text/09_summary}

\section{Acknowledgment}
\input{text/10_acknowledgment}

\section{Code Availability Statement}
\input{text/11_code_availability}

\bibliographystyle{lucas_unsrt}
\bibliography{references}

\end{document}

%% file: text/01_intro.tex
\section{Introduction}

Machine learning (ML) has seen a huge expansion in its range of uses over the last decade.
It is difficult to find a field of industry or science that has not at least explored ML in some capacity.
One particular field where ML usage has seen widespread interest is in high energy physics, which benefits from complex multidimensional problems, large datasets of accurate simulation, and substantial existing computing infrastructure.
These all contribute to a field which has adopted ML algorithms for many aspects of research.
While most ML algorithms in high energy physics are run using CPUs and graphics processing units (GPUs) which provide inference latencies in the milliseconds, field-programmable gate arrays (FPGAs) and application-specific integrated circuits (ASICs) have begun to be used for those applications that demand low latencies~\cite{Duarte:2018ite,DiGuglielmo:2020eqx,Aarrestad:2021zos,Rankin:2020usv}.
Up to now recurrent neural networks (RNNs) have received relatively little attention for these low latency applications, despite their success in many physics tasks and prevalence in the field at large.

RNNs are neural network architectures that treat their inputs as a sequence with a well-defined order. 
RNNs act successively on each entry of the input sequence, utilizing the same set of weights in each step, allowing for RNNs to act on sequences of variable length. 
Most modern RNN applications utilize one of two recurrent layer implementations: long-short term memory layer (LSTMs)~\cite{LSTM} or gated recurrent unit layers (GRUs)~\cite{GRU}. 
Both LSTMs and GRUs contain \textit{forget gates}, which avoid vanishing gradient problems~\cite{vanishing_gradient} and allows for long-distance correlations to be learned by the algorithm. 
These RNN models, often employed to treat time-series signal processing problems, have been successfully employed by physicists to handle different types of data. 
For example, RNN architectures have been utilized to represent hadronic showers (jets) in collider events, aiding on tasks of identifying different jet types by using their constituents to form sequences~\cite{atlas_rnnip, atlas_tauid, sequence_learning}; have been applied to finding interaction vertices in lepton colliders~\cite{rnn_vertexing_ilc}; and have been explored as monitoring tools for the CERN Large Hadron Collider (LHC) superconducting magnets~\cite{WIELGOSZ201740}. 
In general, RNNs have critical applications also outside the realm of collider physics, having been applied, among others, to gravitational waves detection experiments~\cite{rnn_ligo}, to neutrino detectors from nuclear reactors~\cite{rnn_kamland}, and to the reconstruction of quantum dynamics of superconducting qubits~\cite{rnn_qubits}.

In this paper, we focus on particle physics datasets produced by the LHC at CERN~\cite{CERN_as}. 
Collisions within the LHC occur at 40\,MHz, making it impossible to readout and store the entire collisions record. 
To handle this rate, LHC experiments filter out only interesting events through an online selection system called the trigger. 
These systems are usually set in two stages, where the first stage (Level-1 trigger or L1T) needs to operate at 40 MHz with a latency of O(1\,$\mu$s).
The selections performed at these stages need to ensure that interesting events are kept, while discarding common, non-interesting events, which occur several orders of magnitude more frequently than the former. 
Therefore, utilizing complex algorithms such as RNNs is of utmost importance. 

The severe constraints of the trigger prevent the usage of CPUs and GPUs.
Instead, custom low-latency hardware such as FPGAs and ASICs must be deployed to meet the latency requirements and offer the flexibility to adapt to changing conditions.
These devices are also able to take advantage of high parallelism making their designs both efficient and fast.
ML inference in this regime has not seen much support due to its specialized nature, but some tools specifically designed for ultra-low latency inference have emerged~\cite{hls4ml,finn}.
Support in this area has focused primarily on dense and convolutional layers, owing to their versatility and popularity.
In this work we present a generalized and flexible implementation of RNNs written in High Level Synthesis (HLS) for the hls4ml package \cite{hls4ml_github}.
The implementation supports a wide range of RNN sizes, design requirements, and is capable of translating both GRUs and LSTMs trained in the Keras framework~\cite{keras}.
Using three different benchmark neural network models of varying size, we show that ultra-low latency inference can be achieved within the resource limitations of modern FPGAs.
This integration into hls4ml opens the door for much wider usage of RNNs for low latency applications.
While the focus in this work is on applications in physics, we note that there is also a demand for low-power efficient RNNs in industry as well.

%% file: text/02_relatedwork.tex
\section{Related Work}

Previous work in the realm of fast RNN inference on FPGAs has focused largely on millisecond-latency inferences~\cite{Heelan2018FPGAIO,DBLP:journals/corr/ChangMC15,DBLP:journals/corr/LeeHPCSS16,10.1109/ISCA.2018.00012}.
However, for the applications discussed above we are interested primarily in latencies in the microsecond range, or faster.
Some previous work has explored this design space in the context of low-power sparse LSTMs~\cite{DBLP:journals/corr/HanKMHLLXLYWYD16}, small LSTMs for real-time energy reconstruction~\cite{Aad:2021tru}, RNNs for gravitiation-wave experiments~\cite{Que:2021cqo}, and highly quantized RNNs~\cite{finn_rnn}.
In contrast, the work in this paper is focused on general support for both large and small LSTMs and GRUs for problems with a range of latency and device constraints.
The examples we use are largely chosen from the high energy physics domain, but the applicability is by no means limited to this field.
The work is built on top of high level synthesis (HLS) and the hls4ml framework.

HLS tools are designed to simplify the use of FPGAs by automatically transforming algorithms written in C into the register-transfer level (RTL)~\cite{hls}.
There are multiple advantages of these tools. 
One substantial advantage is that they allow users without a knowledge of highly technical Verilog/VHDL langauges to generate effective RTL~\cite{surveyHLS}.
Additionally, they can greatly simplify the effort required in prototyping designs, especially those that are complex.
FPGA manufacturers like Xilinx and Altera have their own HLS compilers for their devices.
There also exist open-source compilers, such as Catapult HLS~\cite{catapulthls2020}.
In this work we use the Vivado/Vitis compiler from Xilinx \cite{vivado,vivado_wp}.

The hls4ml framework is built on top of HLS compilers, and is capable of converting neural network models into fully-ready HLS projects~\cite{hls4ml}.
The details of the HLS design, in particular the resources and latency, can be controlled through multiple tunable parameters in hls4ml.
These are important to allow a flexible design flow that performs well for a wide range of network sizes, architectures, and FPGAs.
They also allow the design to be optimized for the target use case.
hls4ml already has support for MLPs, CNNs, graph NNs, and several other architectures. 
Building on hls4ml allows for models with these architectures to be interfaced with the models in this paper. 
Furthermore, extensive work has been done to hls4ml to ensure that matrix multiplications and other core ML components are optimized. 
Our work uses the hls4ml design flow in order to leverage these existing framework capabilities.

%% file: text/03_impl.tex
\section{Implementation Details}
\label{sec:impl}

Recurrent neural networks at their core are comprised of many standard operations in machine learning.
Our implementation relies on this fact to avoid re-implementing any unnecessary operations in the hls4ml framework.
Taking the LSTM, we see that each state update requires 4 distinct matrix multiplications, given by
\begin{align}
    i_{t}&=\sigma(W_{i}x_{t}+U_{i}h_{t-1}+b_{i})\nonumber\\
    f_{t}&=\sigma(W_{f}x_{t}+U_{f}h_{t-1}+b_{f})\nonumber\\
    o_{t}&=\sigma(W_{o}x_{t}+U_{o}h_{t-1}+b_{o})\nonumber\\
    c_{t}&=\tanh(W_{c}x_{t}+U_{c}h_{t-1}+b_{c})\label{eq:cellstate}
\end{align}
where $W$ and $U$ are the weight matrices (denoted as the kernel and recurrent kernel, respectively), $b$ are the biases for the input gate, forget gate, output gate, and cell state, respectively, $x_t$ is the input at time-step $t$, and $h_{t-1}$ is the computed recurrence value from the previous state.
Each of the operations involving $W$ and $U$ in Eq.~\ref{eq:cellstate} are standard matrix-vector multiplications and the activation functions can be taken directly from the hls4ml library and integrated into the LSTM-specific layer implementations.
The GRU is composed of two gates (update and reset) and a hidden state, but the weights of the kernel and recurrent kernel for the gates are again packaged together and can thus be handled together with one dense layer call each.
The remaining operations to complete a state update for both an LSTM and a GRU are Hadamard products, which is not part of the existing hls4ml library of operations.
In this paper, we implemented an HLS-optimized Hadamard product.
Because many of the operations in our implementation are taken from the existing hls4ml framework, we are able to trivially support the standard tuning knobs for reuse and precision.

\begin{figure}[ht]
\centering
\includegraphics[width=\textwidth]{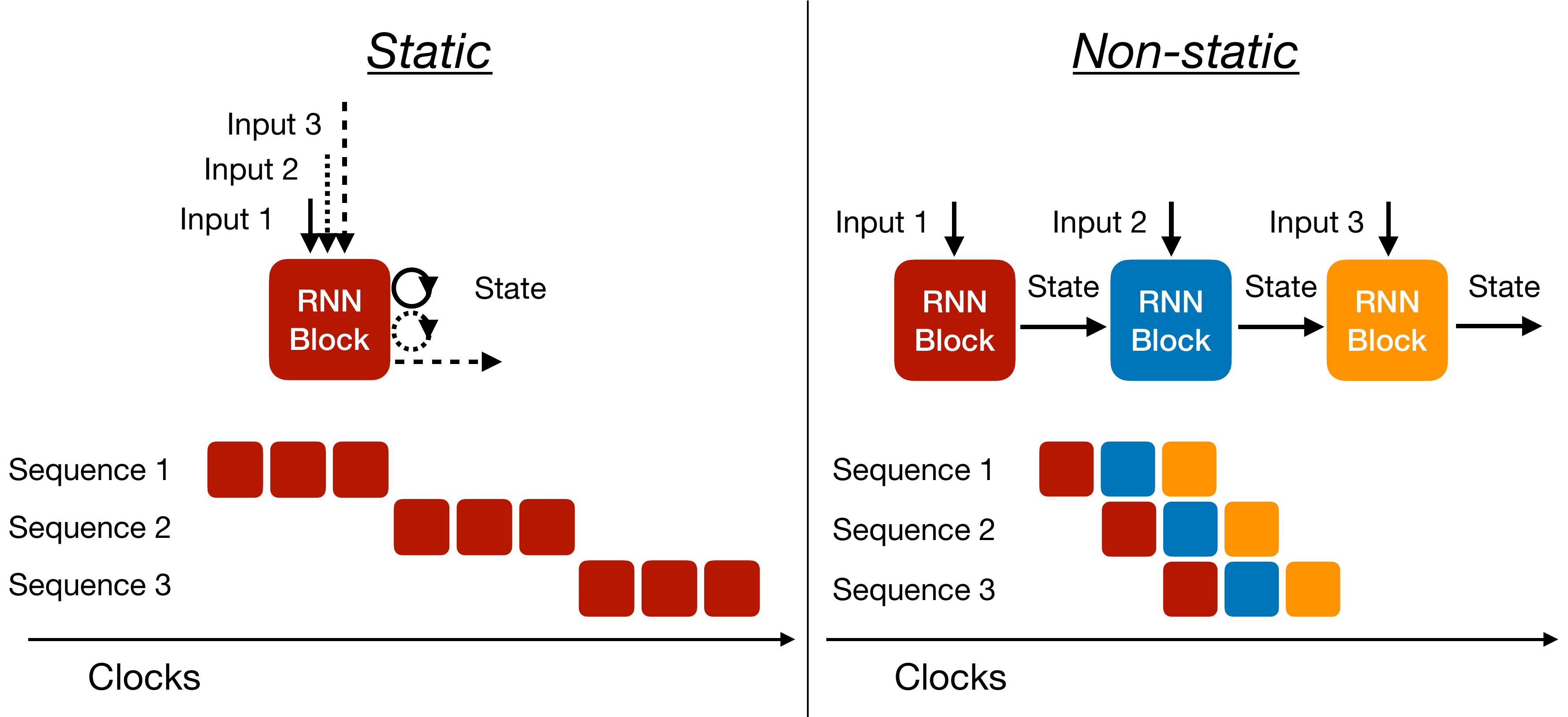}
\caption{Schematics of the two RNN modes: static (left) and non-static (right). A sketch of the latency and pipelining is shown at the bottom for each mode, and illustrates the latency advantage of non-static mode at the cost of resources.}
\label{fig:static_v_nonstatic}
\end{figure}

In addition to the pre-existing hls4ml methods for adjusting resources and latency, RNNs introduce an additional possibility which we refer to as ``static'' and ``non-static'', shown in Fig.~\ref{fig:static_v_nonstatic}.
In static mode, a single RNN block is created, which processes every input for every sequence.
This block stores the necessary state vectors internally, and outputs the final result at the end of the sequence.
Since there is only one RNN block in static mode, the resources are kept to a minimum.
However, the initiation interval (II) of the design increases linearly with the length of the sequence since a new RNN inference cannot begin until the previous inference is complete; in other words, the II is equal to the latency.
In contrast to static mode, non-static mode creates RNN blocks for each input in the sequence, and the necessary state vectors are passed from one RNN block to another.
This results in a resource utilization that is a factor of the sequence length larger than the resource utilization in static mode.
For large RNNs, or inference with long sequences, this means that static mode can be the only viable option.
However, for those RNNs for which it is possible to use non-static mode, the II can be dramatically reduced with respect to static mode since non-static mode allows a new inference to begin once the first RNN block has finished processing the first input from the previous inference.
This reduces the II by a factor of the length of the sequence, and thus increases the overall throughput of the RNN layer by the same factor.
Further increases in throughput are possible when the II of a single RNN block can be made small since in this case each individual block may be used simultaneously by distinct inferences.
While not implemented in this paper, we note that multiple inferences can be cached during static mode when the initiation interval of a single RNN block is less than its latency, thus allowing for higher throughput.

%% file: text/04_benchmarks.tex
\section{Benchmark Studies}
\label{sec:benchmarks}
In order to measure the performance of the hls4ml translation of the RNN-based architectures, we use three different tasks as benchmarks.
The sizes of the networks are chosen such that they span a range of use cases, input dimensions, and numbers of weights.
The first benchmark is a binary classifier with approximately 4000 parameters, trained to classify jets of particles coming from top-quark decays.
The second benchmark is a multi-class classifier, with approximately 50,000 parameters, trained for heavy-flavor jet identification using the reconstructed trajectories of charged particles (tracks) within a jet.
The final benchmark is also a multi-class classifier with approximately 130,000 parameters, trained to classify sequences of strokes into five different image classes.
Each benchmark is studied with two models using LSTM and GRU recurrent layers, respectively. All the models are trained using Keras and TensorFlow.
A summary of each of these models is given in Table~\ref{tab:benchmarks}, and details are given in the following sections.

\begin{table}[ht]
\centering
\caption{Network hyperparameters and total number of trainable parameters for different benchmark models.\label{tab:benchmarks}}
\vspace{0.2 cm}
\begin{tabular}{l|c|c|c|c|c|r|r|r}
\toprule
\multirow{3}{*}{Benchmark} & Sequence & Input & Hidden & Dense & output & \multicolumn{3}{c}{Trainable parameters}  \\
\cmidrule{7-9}
& length & vector & vector & layer & vector & Non-RNN  & LSTM & GRU  \\
&  & size & size & sizes & size & layers &  &   \\
\midrule
Top tagging    & 20   & 6 & 20  & 64      & 1 &  1,409  &  2,160  & 1,680  \\ 
Flavor tagging & 15   & 6 & 120 & 50/10   & 3 &  6,593  &  60,960 & 46,080  \\ 
QuickDraw      & 100  & 3 & 128 & 256/128 & 5 & 66,565  &  67,584 & 51,072  \\
\bottomrule
\end{tabular}
\end{table}

%% file: text/05_top.tex
\subsection{Top Quark Tagging}
The top quark tagging algorithm is trained to classify top quarks from light-flavor quarks using simulated events generated at $\sqrt{s} = 13\TeV$ for comparison to LHC performance. 
Algorithms designed for this task could be utilized in the Level-1 trigger systems of LHC experiments to help increase the acceptance for these types of interesting decays.
Their use would require algorithm latencies of less than approximately a few microseconds in order to fit within system constraints.

The data used for training and testing consists of parton-level scattering processes with top quark-antiquark pair ($\mathrm{t}\overline{\mathrm{t}}$) and light-flavor quark-antiquark pair ($\mathrm{q}\overline{\mathrm{q}}$) final states that are generated at leading-order using MadGraph~\cite{aMCNLO} with the NNPDF23LO1 parton distribution functions (PDFs)~\cite{NNPDF23LO1}. 
The transverse momenta (\pt) of the partons are generated in a window with energy spread given by $\delta \pt /\pt = 0.01$, centered at 1\TeV. 
These parton-level events are then decayed and showered using Pythia8~\cite{SJOSTRAND2015159} (version 8.212) with the Monash 2013 tune \cite{monash-tune}, including the contribution from the underlying event. 
Jets are clustered using the anti-\kt algorithm, with a distance parameter of $0.8$. 
Only low level features are used in this study. Particles inside a jet are ordered according to their \pt and up to 20 particles are used in this study. 
For each particle six features are use: $\pt$, pseudorapidity ($\eta$), azimuthal angle ($\phi$), energy, relative angular distance from the jet axis, particle ID given by the generator. 
The generated events are split into training (95\%) and testing (5\%), and during training 20\% of the training data is used for validation. 

Two networks are trained for identifying the jets coming from top-quark decay. 
The padded sequence of particles, with maximum length of 20, is fed into a recurrent layer with an output size of 20. 
The output from the final recurrent layer is passed through a dense layer of size 64 before sending it to the output layer. 
The recurrent layer used sigmoid and hyperbolic tangent activation functions. 
The activation function for the hidden layers is ReLU \cite{nair2010rectified} while the output layer activation function is a sigmoid function. 
The binary cross-entropy loss function is minimized with L1 ($10^{-5}$) and L2 ($10^{-4}$) regularization of the weights  using the Adam algorithm~\cite{DBLP:journals/corr/KingmaB14} with a learning rate of 2$\times10^{-4}$ and a batch size of 246. 
The two models use different recurrent layers; one uses GRU and the other uses LSTM. There are total 3,089 and 3,569 trainable parameters for the LSTM and GRU models, respectively.


%% file: text/06_ftag.tex
\subsection{Jet Flavor Tagging}

The jet flavor tagging algorithm was trained on CMS Open Data samples containing top quark pairs decaying hadronically with center-of-mass energy of 7\TeV \cite{CERN_opendata}. 
These events are rich in bottom quark jets (b jets), charm quark jets (c jets) and jets from light quarks and gluons (light jets), and so are optimal for training this class of algorithms. 
Jets are labeled b jets if they contain bottom quarks, c jets if they do not contain bottom quarks but contain charm quarks, and light jets if they do not contain bottom or charm quarks. 
The main feature that separates b jets (and c jets) from light jets is the presence of the displaced vertex corresponding to the decay of the hadron containing the b (or c) quark. 
These hadrons can travel significant distances before decaying due to their mass, depending on their momenta. 
The algorithm proposed here aims to identify the presence of tracks that are consistent with these displaced vertices with the usage of an RNN architecture. 
This strategy is inspired by the RNNIP algorithm used by the ATLAS experiment~\cite{atlas_rnnip}.

In this study, we consider jets reconstructed with the anti-kt algorithm with a distance parameter of $0.5$, with transverse momenta larger than 30\GeV and absolute pseudorapidities less than 2.0. 
Tracks with transverse momenta larger than 1\GeV are associated to the nearest jet according to the angular distance $\Delta R$; a maximum $\Delta R$ of 0.5 is required for association. 
Tracks within a jet are ordered by the significance of their transverse impact parameter ($\mathcal{S}(d_0)$), and only the first 15 tracks are used by the algorithm. 
Each track is represented by a vector of the following features: relative transverse momentum ($\pt(\mathrm{track})/\pt(\mathrm{jet})$); $\Delta R(\mathrm{track, jet})$; transverse and longitudinal impact parameters ($d_0$, $d_z$) and their significances ($\mathcal{S}(d_0)$, $\mathcal{S}(d_z)$).

Flavor tagging models were constructed using Keras/TensorFlow, using either GRU or LSTM layers. 
The padded sequence of tracks, with maximum length of 15, is fed into either one recurrent layer (GRU or LSTM) with 120 hidden units. 
The recurrent layer outputs a representation of the padded sequence directly into two dense layers with ReLU activation function, with 50 and 10 hidden layers. 
The following layer outputs the probabilities of a jet to be classified as either a b jet, c jet or light jet; it contains three output nodes with softmax activation function. 
The training is performed with a categorical cross-entropy loss, with 30\% of the training data retained as the validation dataset and used for early stopping based on the accuracy metric. 
The GRU (LSTM) architecture contains 52,673 (67,553) trainable parameters, of which 46,080 (60,960) are in the recurrent layer.

%% file: text/07_qdraw.tex
\subsection{Quickdraw Dataset}

The QuickDraw dataset~\cite{qdraw_dataset} is a collection of 50 million drawings in 345 categories created by Google and contributed by players of the game Quick! Draw. 
In this game, users are asked to draw a specified drawing in under 15 seconds. 
The drawings are recorded as a time-stamped sequence of the strokes from which the drawing is created.
For each 15 second stroke the $x$ and $y$ coordinates of the pen are recorded 100 times.
The coordinates along with the timestamp make up the network inputs.
While 345 different drawing categories exist, we use only 5 for our tests; these are \textit{ants}, \textit{butterflies}, \textit{bees}, \textit{mosquitos} and \textit{snails}.
Contrary to other popular representations of images, the QuickDraw dataset is completely stroke-based. 
We train two recurrent neural networks to classify images into each respective category.
Similar algorithms could be used to identify tracks based on the sequence of their hits, or incident particles based on their showers in finely-segmented calorimeters.
While not appropriate for the Level-1 trigger environment, running these algorithms at later stages in the trigger for a whole event could still require low latencies under a millisecond depending on the exact application.


The two networks process the sequence of 100 stroke inputs from the QuickDraw dataset with a recurrent layer whose output size is 128.
The final recurrent layer output is passed through two dense layers of sizes 256 and 128, respectively, before being sent to the final output layer.
Dropout layers are placed before the two dense layers to regularize during training.
The recurrent layer uses a hyperbolic tangent activation function, the dense layers use ReLU activations, and the output is a five-class softmax layer.
The only difference in the two networks is that in one the recurrent layer is a GRU and in the other it is an LSTM.
These networks have total sizes of 117,637 and 134,149 trainable parameters, respectively.
The two networks perform well and show top-1 AUCs that are nearly identical, approximately 99\% for each of the five classes.

%% file: text/08_results.tex
\section{Performance, Resource and Latency Estimation}

The models described in Sec.~\ref{sec:benchmarks} are translated into HLS using the hls4ml framework.  
Vivado HLS 2019.2 is used for HLS synthesis with the synthesis clock frequency set to 200\,MHz. 
For the top tagging and flavor tagging models we use a Xilinx Kintex UltraScale FPGA (part number \texttt{xcku115-flvb2104-2-i}) as the target device and for the Quickdraw models we use a Xilinx Alveo U250 (part number \texttt{xcu250-figd2104-2-e}).
For each of the three benchmark models a range of different settings are considered simultaneously to accurately profile the full design space. 
These settings modify the quantization of the model by adjusting the fixed-point data type and modify the degree of parallelism of the design by using the hls4ml reuse parameter.
Finally, we also consider the static and non-static implementations discussed in Sec~\ref{sec:impl}.

\subsection{Quantization}

The weights and biases in trained models are typically stored with 32-bit floating-point precision. 
However, 32-bit floating-point calculations are often not required for optimal network inference, and are costly to implement on FPGAs. 
Other quantization techniques can offer more efficient ways of compressing neural networks by reducing the number of bits used to represent the weights and biases, ideally with no or minimal loss in performance. 
In hls4ml, all the inputs, weights, biases, sums, and outputs of each layer are represented as fixed-point numbers.
In this scheme the amount of bits used to store the integer and decimal components of the number are configured, such that, for example, an unsigned fixed point number with 4 integer bits and 3 fractional points is capable of storing values between 0 and 15.875 with a granularity of 0.125.
The total numbers of bits is also referred to as the precision of the fixed-point number.
Hls4ml allows a different precision to be chosen for the computations and internal values of each individual layer; for the sake of consistency we fix the precision to be the same for all layers in the scans below.
We do find that it is necessary to increase the precision and size of the LUT used for the softmax computation at the end of the flavor-tagging and QuickDraw models, but this has a minimal impact on the overall resource usage.

\begin{figure}[ht!]
\centering

    \begin{subfigure}[b]{0.33\textwidth}
    \centering
    \includegraphics[width=\textwidth]{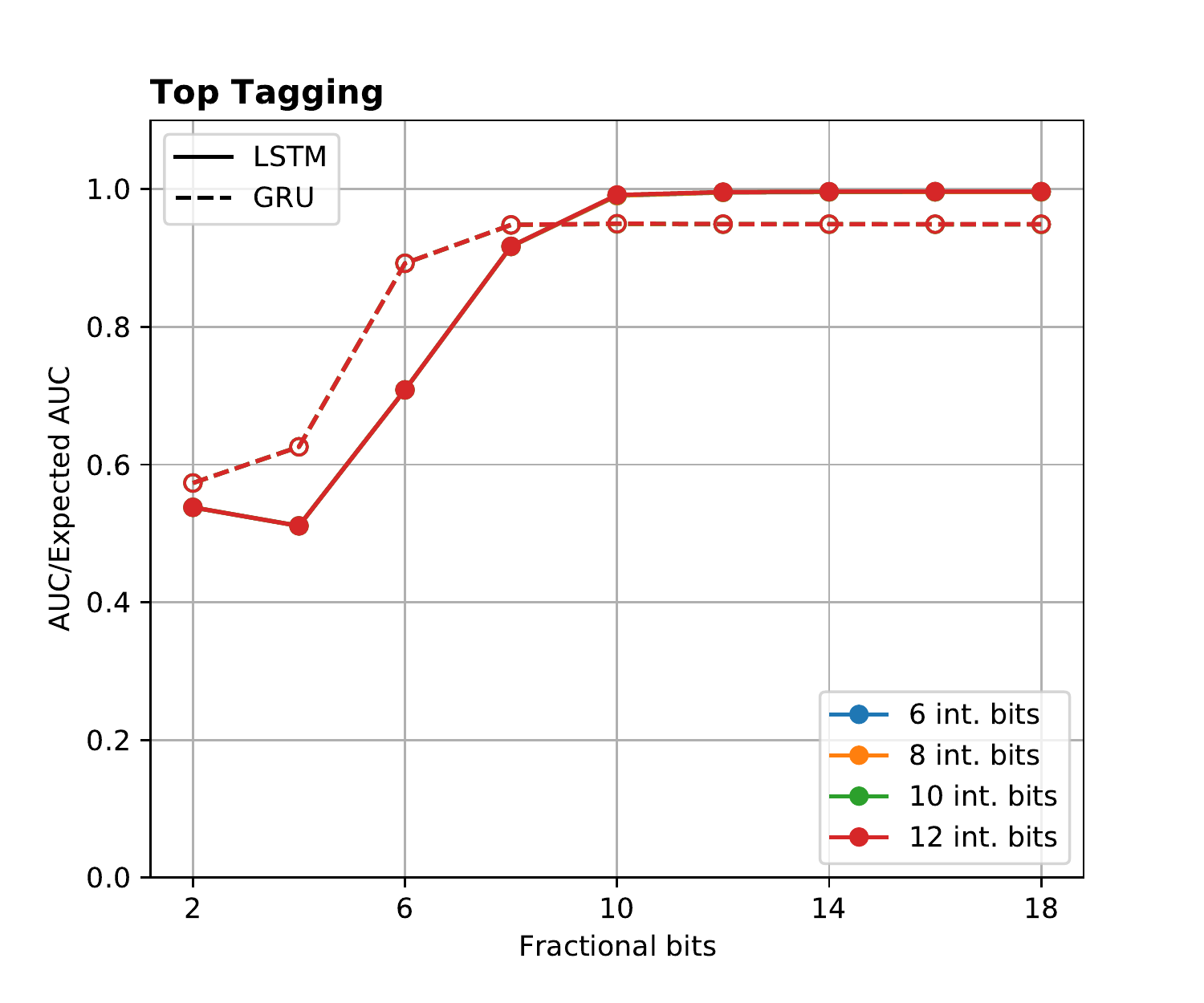}
    \caption{Top quark tagging}
    \end{subfigure}
    \begin{subfigure}[b]{0.33\textwidth}
    \centering
    \includegraphics[width=\textwidth]{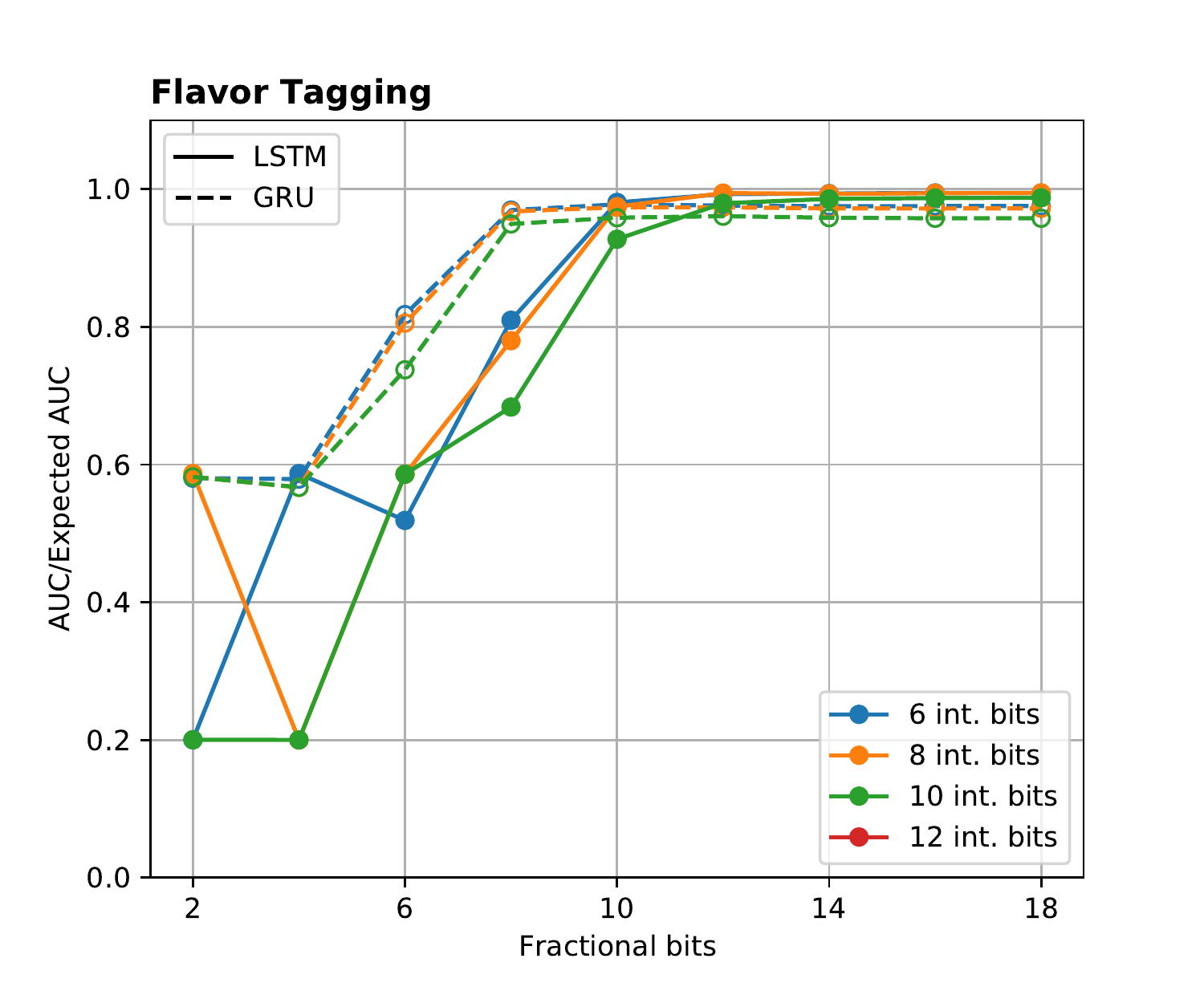}
    \caption{Jet flavor tagging}
    \end{subfigure}
    \begin{subfigure}[b]{0.33\textwidth}
    \centering
    \includegraphics[width=\textwidth]{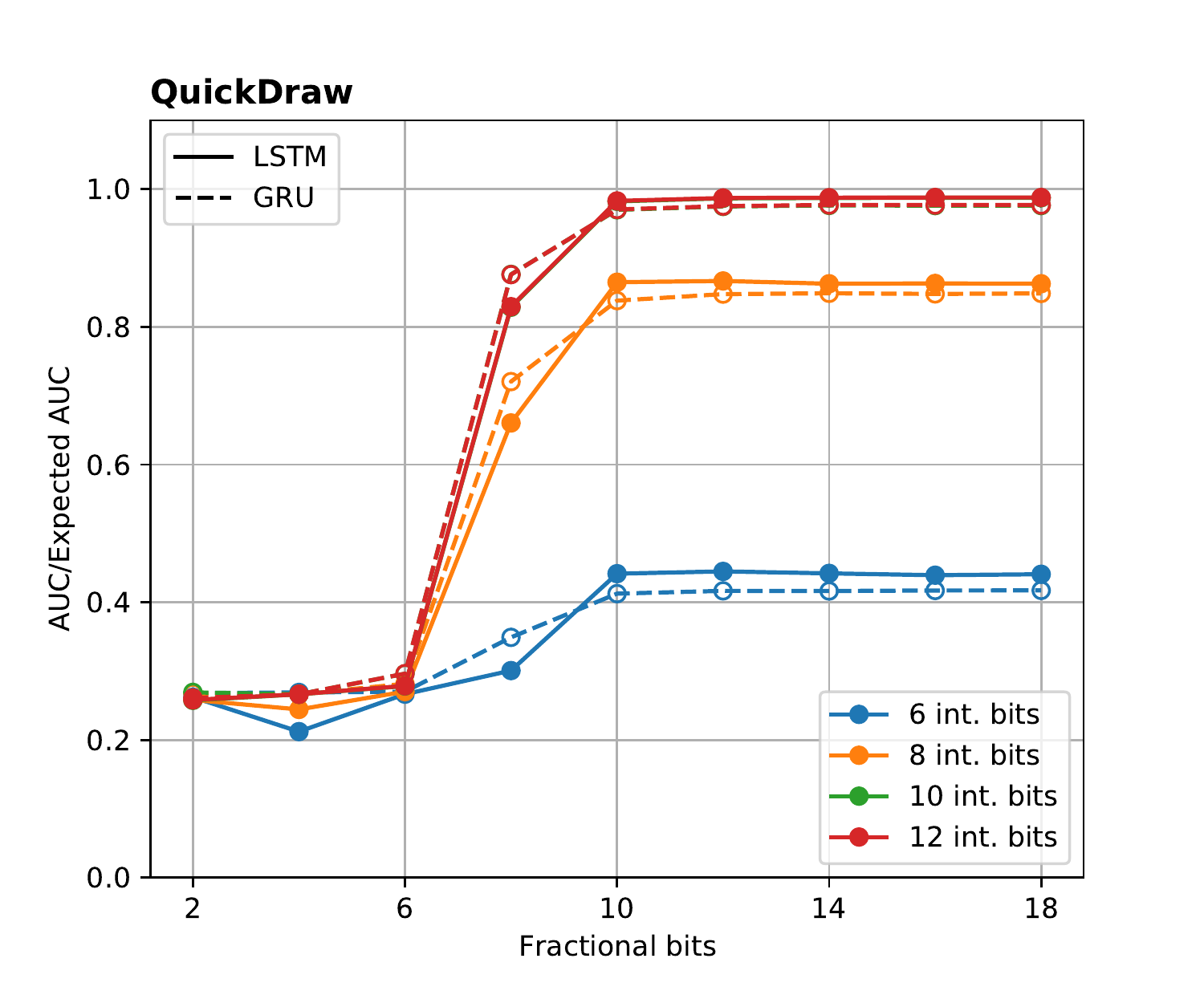}
    \caption{QuickDraw}
    \end{subfigure}
    
\caption{Ratios of the fixed-point and floating-point AUCs as function of fractional bits for the (a) top quark tagging, (b) jet flavor tagging, and (c) QuickDraw models. 
The precision of the integer part is kept fixed to 6 (blue), 8 (orange), 10 (green), and 12 (red) bits.
All four lines overlap for the top-tagging GRU (dashed) and LSTM (solid) models.}
\label{fig:relauc}
\end{figure}

The optimal precision for each model depends on the training details, the specific task, and the inputs. All the models are quantized only after training, a method referred to as ``post-training quantization'' (PTQ).
For each model we profile the performance of the synthesized design from hls4ml as a function of the bit precision of the weights and activation functions. 
Figure~\ref{fig:relauc} shows the ratio of the AUC from the quantized model to the AUC from the floating-point model as a function of fractional bits while keeping the precision of the integer part fixed to 6, 8, 10, or 12 bits.

The best performance for each model, measured by the AUC ratio, is generally achieved with at least 10 fractional bits irrespective of the values of the integer bit. 
For the top quark and flavor tagging models, 6 integer bits are sufficient, while the QuickDraw models require at least 10 integer bits. 
For further results in this paper, we fix the integer bits for each model to these values. 
We note that there is a small performance degradation in the GRU models after quantization for all three benchmark cases. 
The difference is particularly visible for the Top Tagging model, but it is less than 5\%. 
It is possible that quantization-aware training or sequence masking could potentially enable models with lower precision to perform as well as or better than the ones we present in this paper.


\subsection{Parallelization}

\begin{figure}[t!]
\centering

    \begin{subfigure}[b]{0.33\textwidth}
    \centering
    \includegraphics[width=\textwidth]{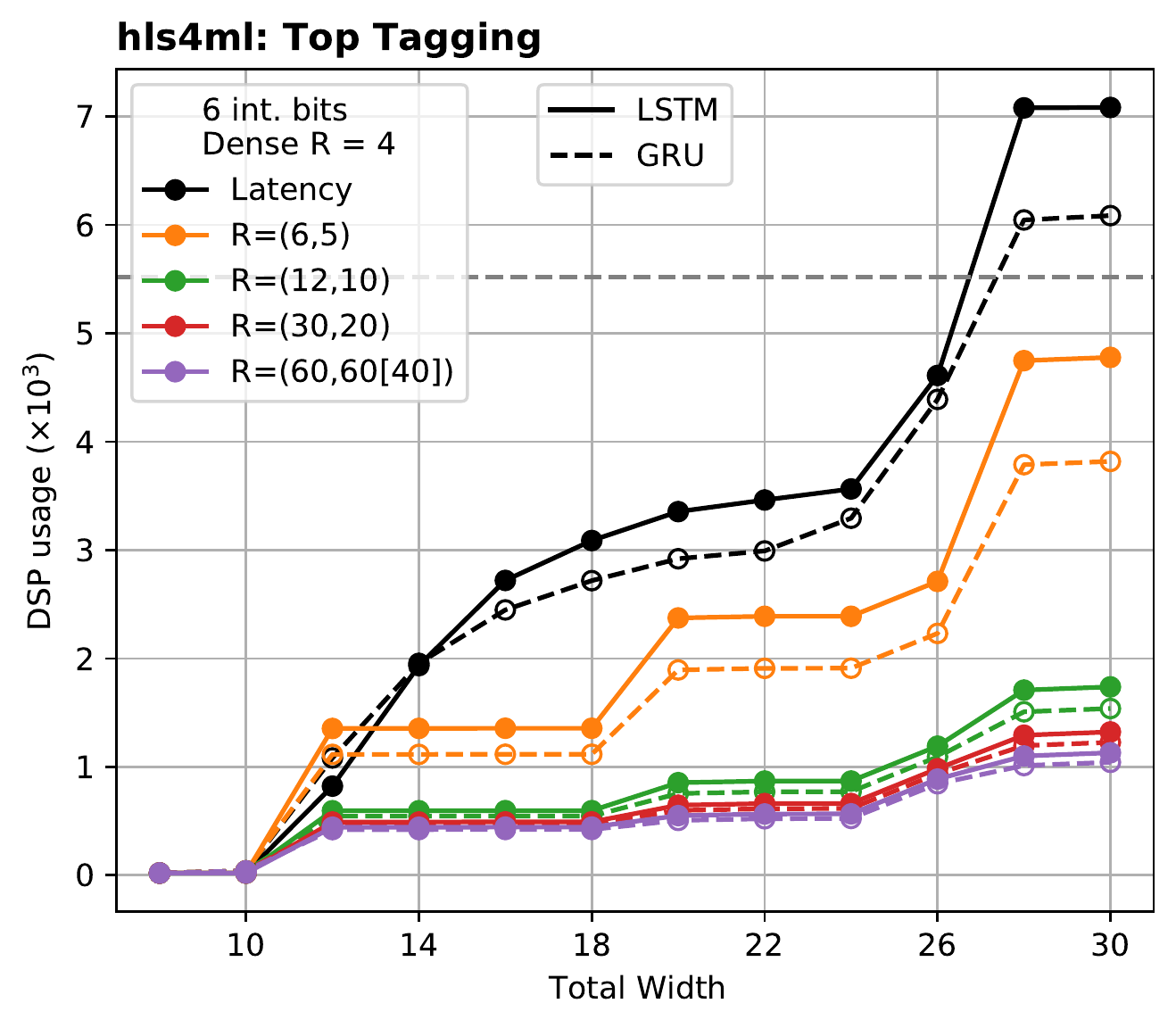}
    \caption{Top quark tagging}
    \end{subfigure}
    \begin{subfigure}[b]{0.33\textwidth}
    \centering
    \includegraphics[width=\textwidth]{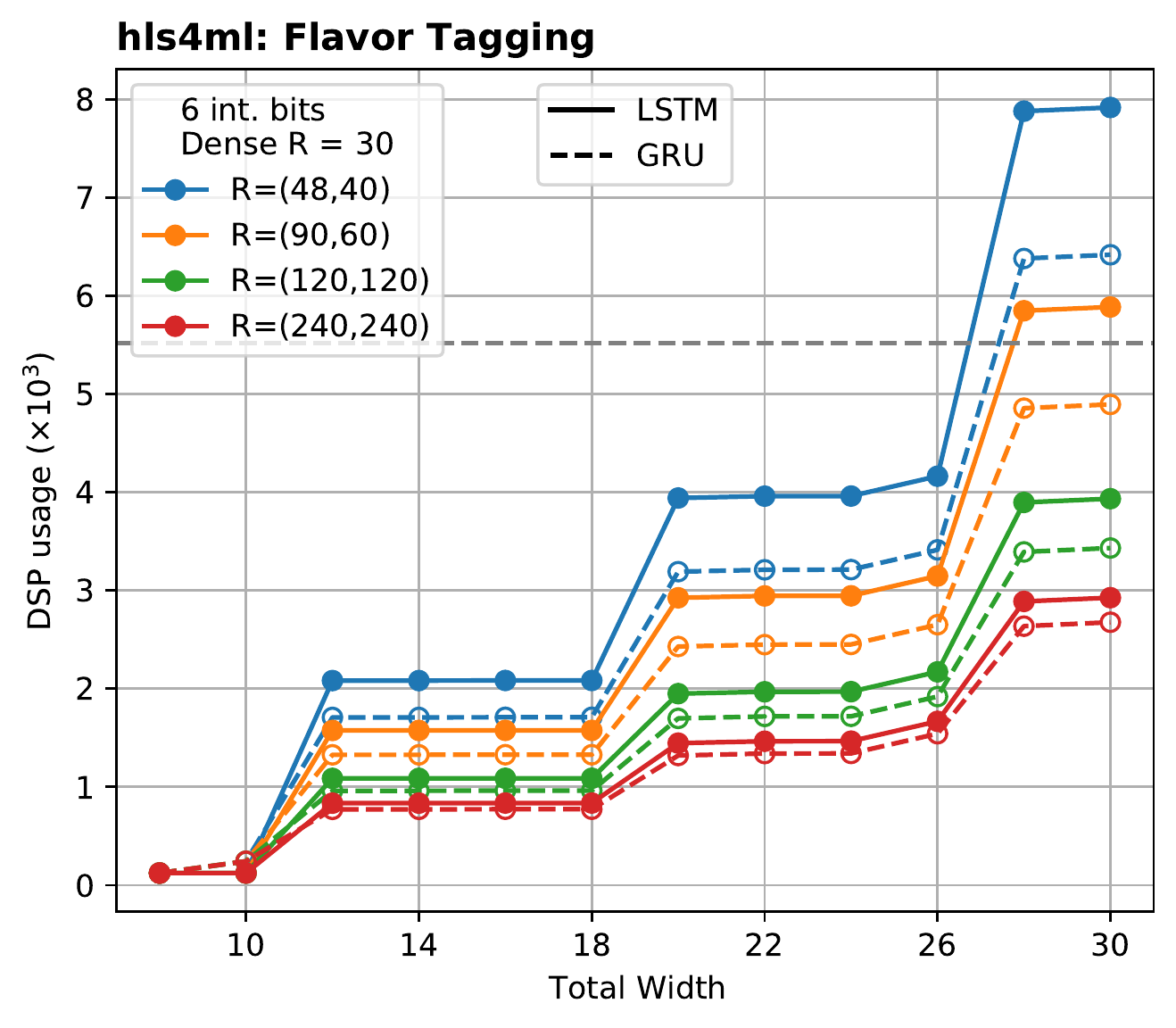}
    \caption{Jet flavor tagging}
    \end{subfigure}
    \begin{subfigure}[b]{0.33\textwidth}
    \centering
    \includegraphics[width=\textwidth]{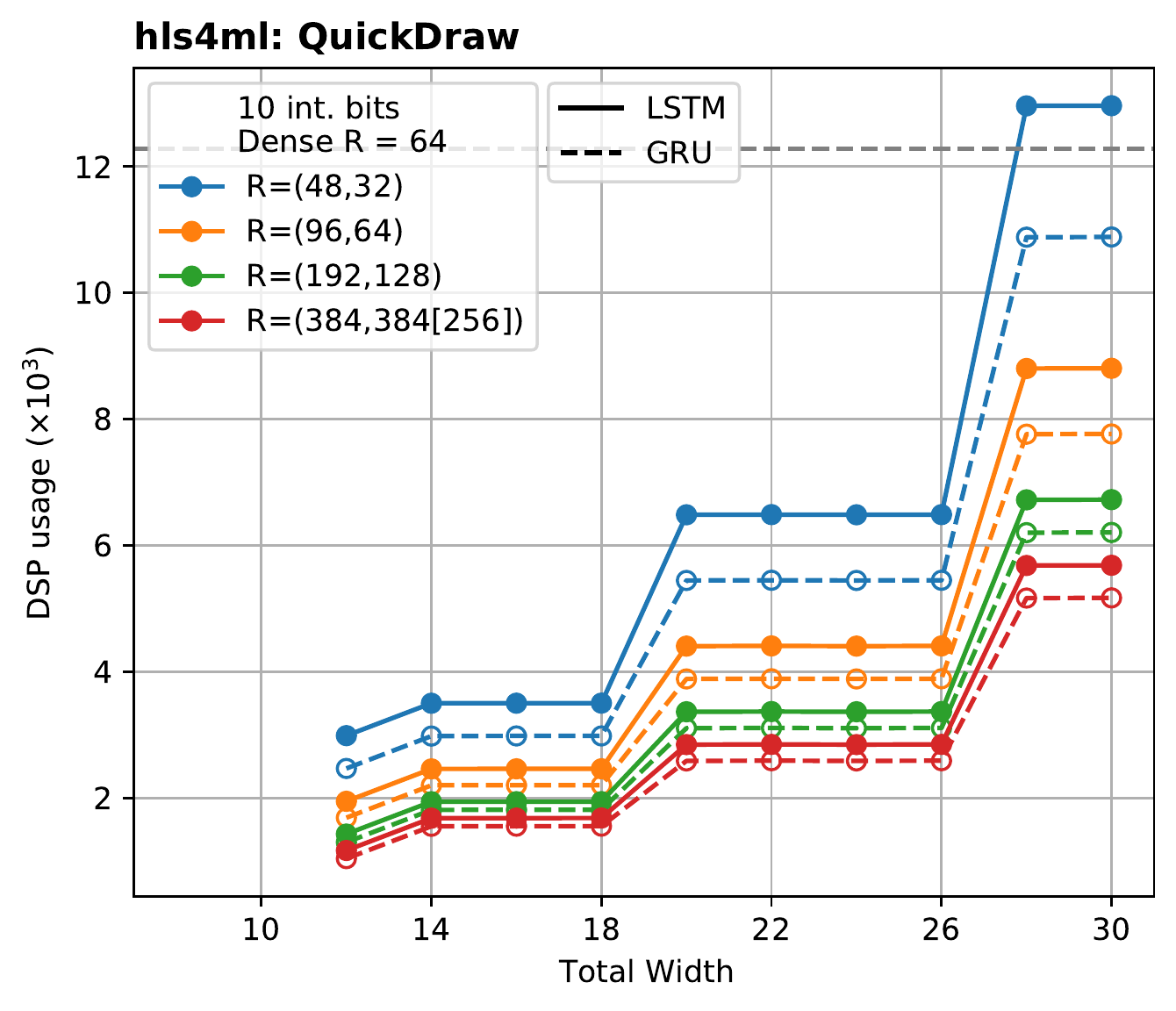}
    \caption{QuickDraw}
    \end{subfigure}
    
\caption{DSP utilization as a function of total width for the (a) top quark tagging, (b) jet flavor tagging (c) QuickDraw model. 
Performance of both GRU (dashed) and LSTM (solid) models are shown. 
DSP utilization with latency strategy is shown only for the top quark tagging models and all other lines correspond to different reuse factors. 
The DSPs available in the target FPGA for each model are shown in the black dashed horizontal line.}
\label{fig:util_dsp}
\end{figure}

\begin{figure}[ht!]
\centering

    \begin{subfigure}[b]{0.33\textwidth}
    \centering
    \includegraphics[width=\textwidth]{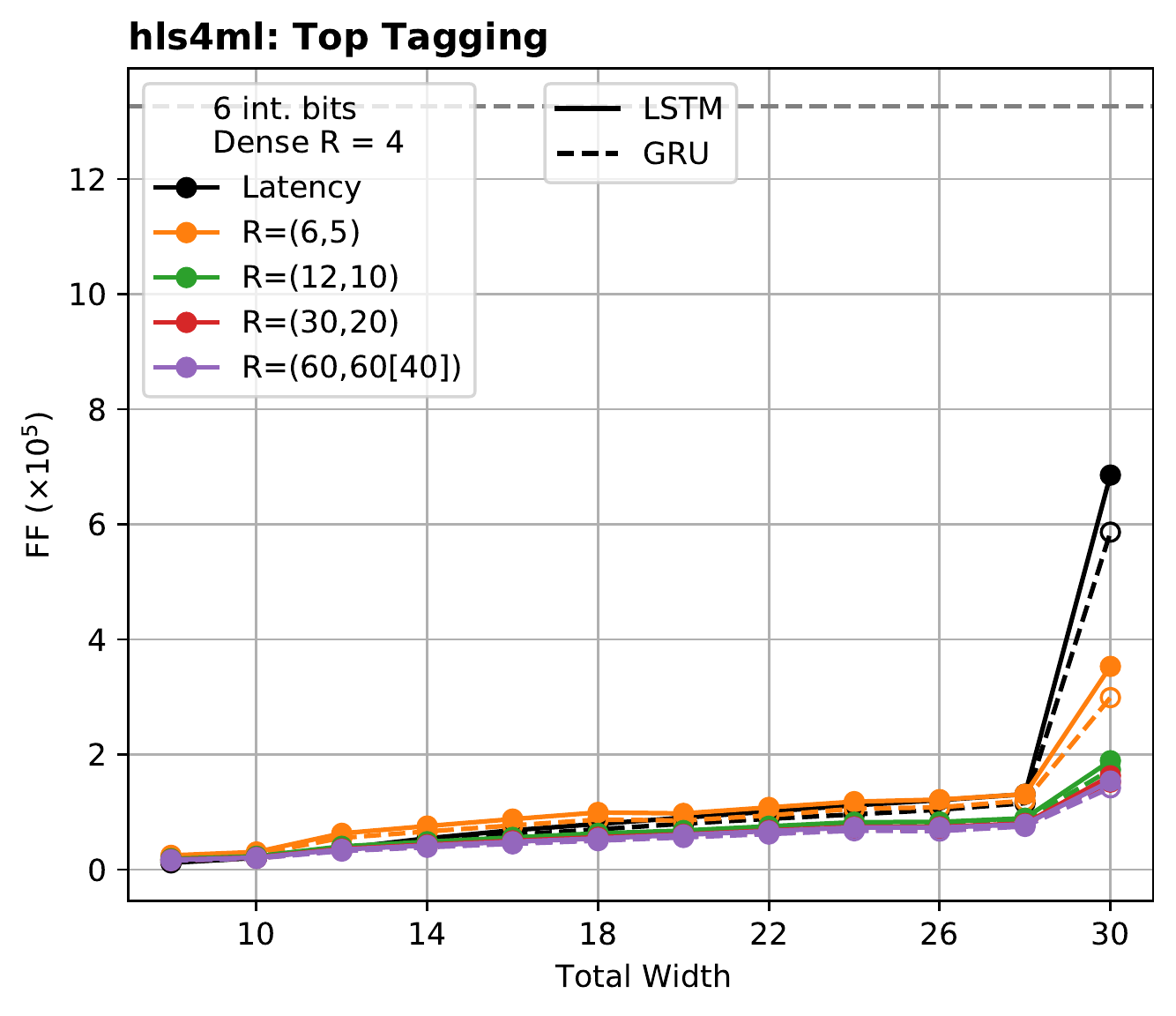}
    \caption{Top quark tagging}
    \end{subfigure}
    \begin{subfigure}[b]{0.33\textwidth}
    \centering
    \includegraphics[width=\textwidth]{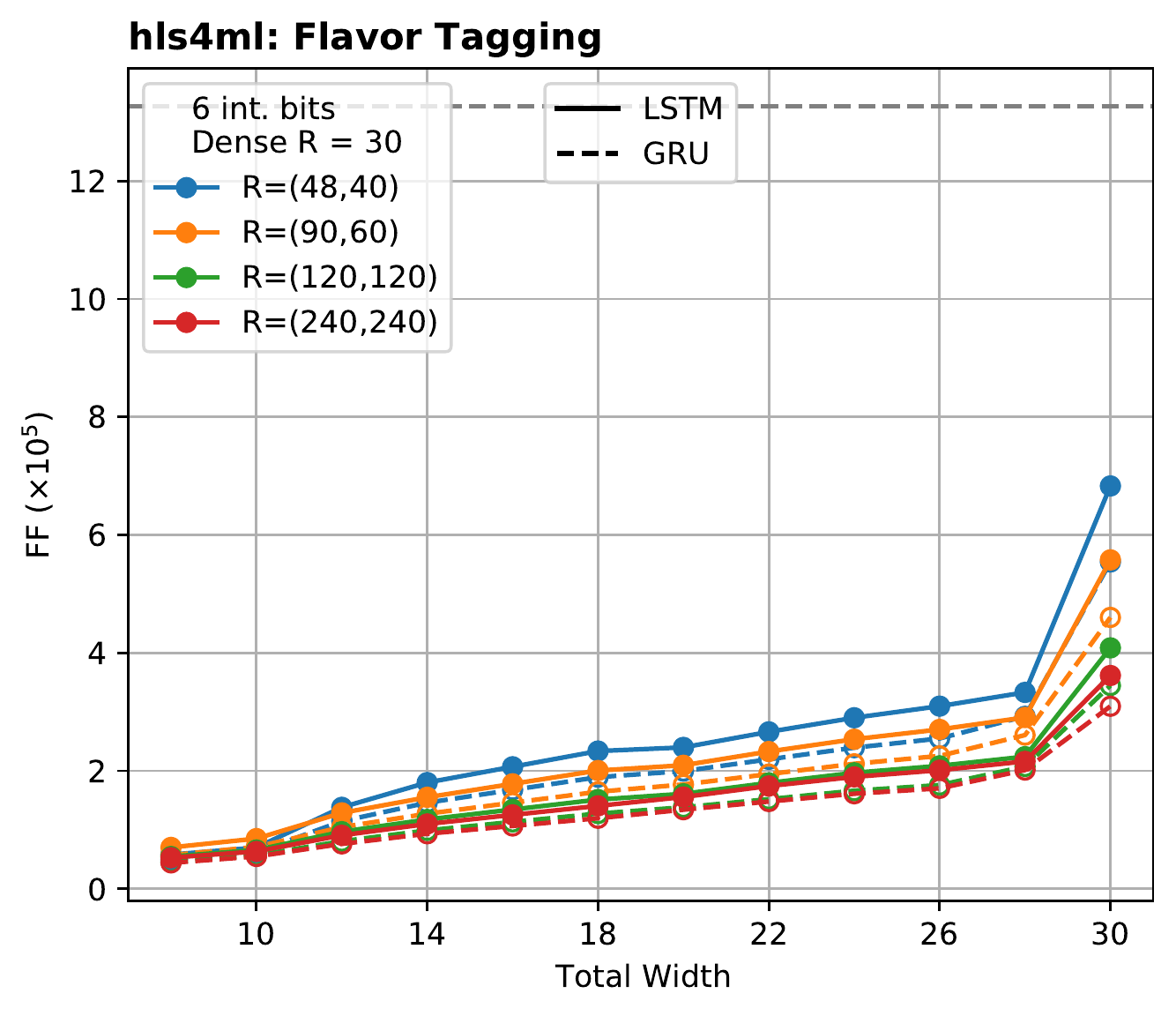}
    \caption{Jet flavor tagging}
    \end{subfigure}
    \begin{subfigure}[b]{0.33\textwidth}
    \centering
    \includegraphics[width=\textwidth]{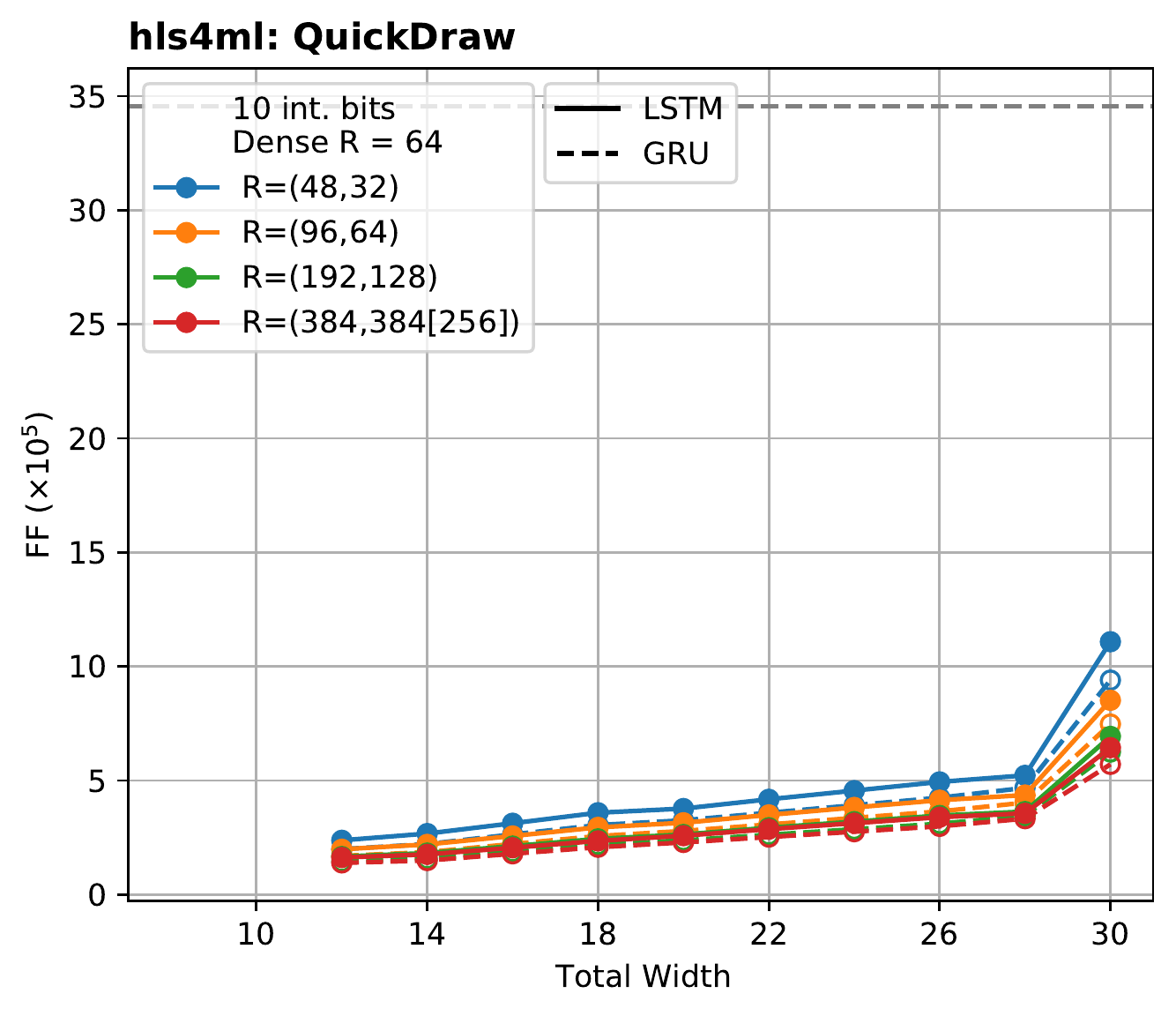}
    \caption{QuickDraw}
    \end{subfigure}
    
\caption{FF utilization as a function of total width for the (a) top quark tagging, (b) jet flavor tagging (c) QuickDraw model. 
Performance of both GRU (dashed) and LSTM (solid) models are shown. 
FF utilization with latency strategy is shown only for the Top Tagging models and all other lines correspond to different reuse factors. 
The FFs available in the target FPGA for each model are shown in the black dashed horizontal line.}
\label{fig:util_ff}
\end{figure}

\begin{figure}[h!]
\centering

    \begin{subfigure}[b]{0.33\textwidth}
    \centering
    \includegraphics[width=\textwidth]{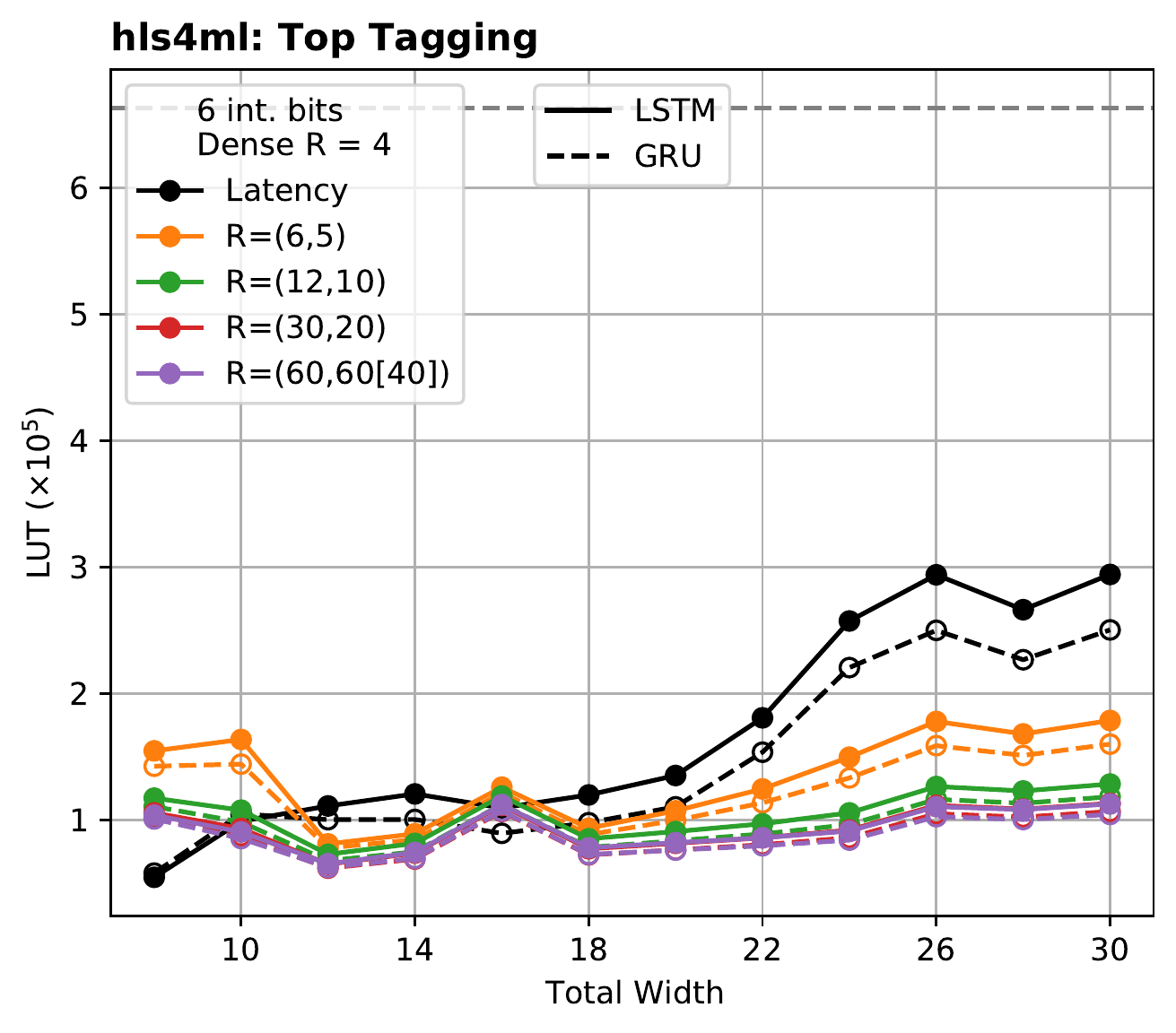}
    \caption{Top quark tagging}
    \end{subfigure}
    \begin{subfigure}[b]{0.33\textwidth}
    \centering
    \includegraphics[width=\textwidth]{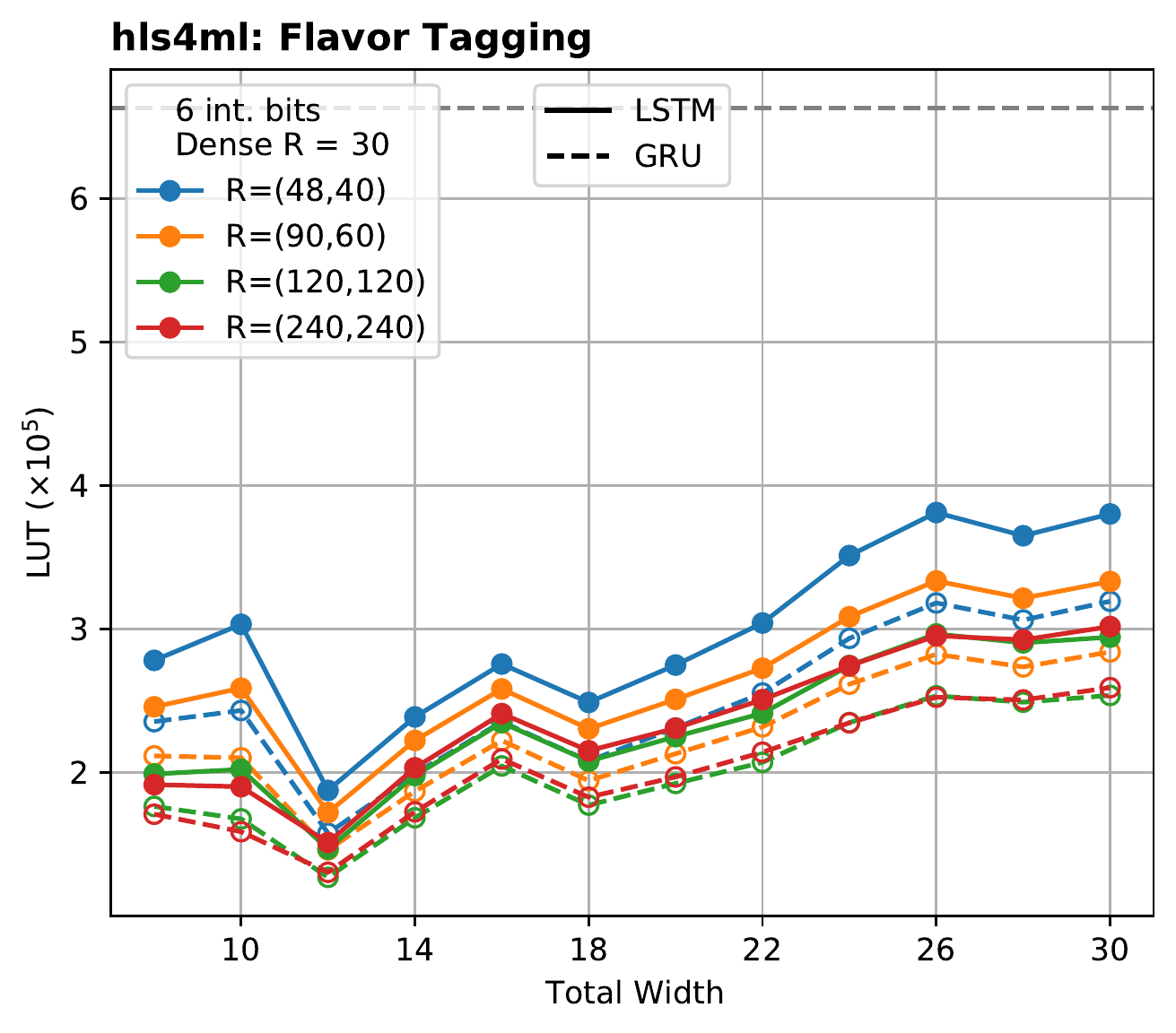}
    \caption{Jet flavor tagging}
    \end{subfigure}
    \begin{subfigure}[b]{0.33\textwidth}
    \centering
    \includegraphics[width=\textwidth]{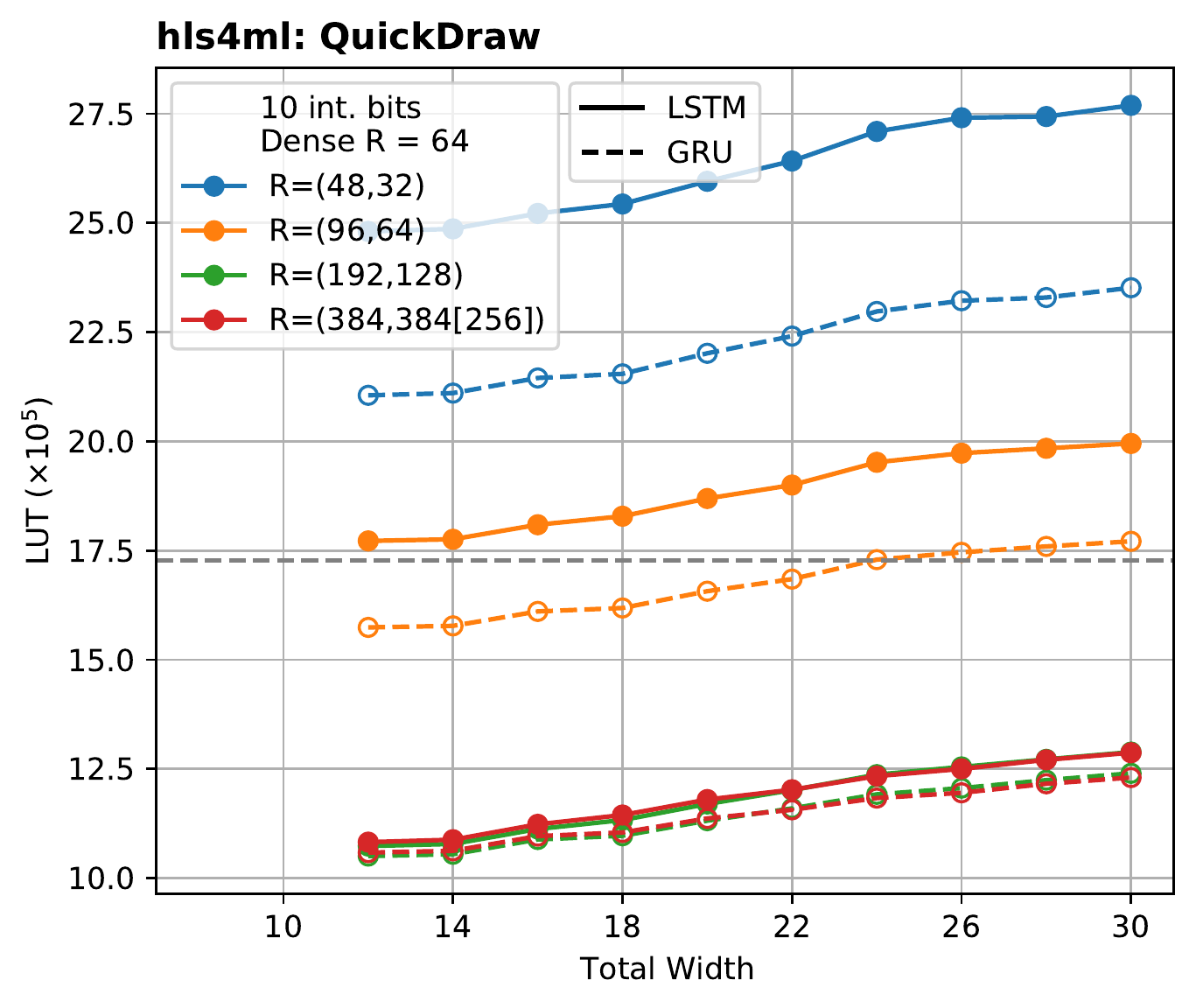}
    \caption{QuickDraw}
    \end{subfigure}
    
\caption{LUT utilization as a function of total width for the (a) top quark tagging, (b) jet flavor tagging (c) QuickDraw model. 
Performance of both GRU (dashed) and LSTM (solid) models are shown. 
LUT utilization with latency strategy is shown only for the Top Tagging models and all other lines correspond to different reuse factors. 
The LUTs available in the target FPGA for each model are shown in the black dashed horizontal line.}
\label{fig:util_lut}
\end{figure}

The other main tuning knob besides the precision is the amount of parallelism employed during weight matrix multiplication.
This is controlled in hls4ml through a parameter called ``reuse''.
Specifically, reuse is the number of multiplication operations each digital signal processing (DSP) block must do for a given matrix multiply.
Setting reuse to 1, i.e. the fully parallel case, means that each multiplication is done by its own DSP and can happen simultaneously.
Increasing the reuse factor reduces the number of DSPs that are required, but increases the latency and initiation interval of the layer computation in proportion to the reuse.
All three benchmark models are synthesized with different values of the reuse factor ($R$) and fractional bit precision.
The results are expressed for different FPGA resource categories: onboard FPGA memory (BRAM), DSPs, and registers and programmable logic like flip-flops (FFs) and lookup tables (LUTs). 
In hls4ml, a model can either be synthesized to minimize the latency (\textit{latency strategy}) or the resource utilization (\textit{resource strategy}). 
For large models with ~40k or more trainable parameters it becomes difficult to synthesize the models with the latency strategy, and so resource strategy must be used.
With resource strategy the design is optimized for low resource utilization by reusing existing hardware to complete operations in multiple stages. 
Out of the three benchmark models only the Top Tagging model is small enough to be synthesized with both latency and resource strategies, whereas only resource strategy is used for the other two models. 
The minimum and maximum latencies for each model are shown in Table~\ref{tab:toptag},~\ref{tab:flavortag}, and~\ref{tab:quickdraw}.
The amounts of DSPs, FFs, and LUTs for each model are shown for different $R$ values in Fig.~\ref{fig:util_dsp},~\ref{fig:util_ff}, and~\ref{fig:util_lut}, respectively.
In these results the reuse factor values are written in the form $R= (X, Y)$, where $X$ and $Y$ correspond to the reuse factors for the kernel and recurrent kernel matrix multiplications discussed in Eq.~\ref{eq:cellstate}.
The the numbers shown in the square brackets correspond to reuse factor for the LSTM layer in the case when the reuse factor differs between the LSTM and GRU implementations.

We observe that all resources generally increase with smaller values of $R$ and increased precision.
In the case of FFs and LUTs this increase is roughly linear, while for DSPs the utilization remains flat until the precision exceeds the DSP input width.
The latency, on the other hand, follows a scaling inverse to that of the FFs and LUTs with respect to the reuse.
Thus, as with other architectures supported under hls4ml, reuse can be used to reduce FFs and LUTs at the expense of latency.
This simple scaling is critical for allowing users to tune the resource usage and latency such that the synthesized designs to meet desired requirements.
The latency strategy adds another finer option to this tuning space for latency-limited tasks, but comes at the cost of larger resource usage.
As expected we find that the GRU models use approximately $1/4$ less resources when compared to the LSTM models.
This is a result of the 3:4 ratio between the number of matrix multiplications in GRU and LSTM models.
Finally, it is important to note that the results shown in this paper are from HLS synthesis.
When running Vivado synthesis we observe a reduction in LUT usage between 20\% and 65\% and in FF usage between 10\% and 20\%.
This is particularly important to note in the case of the larger flavor tagging and QuickDraw models where the estimated LUT usages from HLS synthesis are quite large.

For the top tagging models we observe that designs with maximal quantized performance can be implemented on one SLR of a Xilinx Virtex Ultrascale+ VU9P, the planned future device for an upgrade to the CMS Level-1 trigger~\cite{CMSP2L1T}.
We observe slightly larger resource usage for the flavor tagging models, as expected, but still within the resource constraints of a single SLR of a VU9P.
In both cases the latencies for the designs are also within the task requirements.
For the QuickDraw models, the estimates from Vivado synthesis suggest that maximal quantized performance could be implemented on a Xilinx Alveo U250, a popular device for the types of coprocessor applications we envision for these models.
Extrapolating from the II, we find the average throughput of the QuickDraw LSTM model is between 4300 to 9700 events/sec.
Tests of the batch 1 inference for the same model using an Nvidia Tesla V100 GPU yield a throughput of 660 events/sec. 
Increasing the batch size to 10 increases the throughput to 7700 events/sec, comparable with the FPGA throughput. 
The throughput increases further by a factor of five to approximately 30000 if the batch size is increased to 100.
While it is unsurprising that GPU inference at large batch sizes is able to outperform an FPGA, many physics tasks are inherently low-batch problems.
This is because each event must be processed separately and latency is extremely important, therefore the maximal batch size is dictated by the amount of inferences necessary only for a single event.
For example, algorithms to classify particle-induced showers in a detector need only to be run once every event if the algorithm can use the full detector information in one pass.
Thus, a factor of ten improvement in the FPGA inference for batch-1 inference is highly relevant for future trigger applications.

\begin{table}[ht!]
\centering
\caption{ Minimum and maximum latencies for the top quark tagging model.}
\label{tab:toptag}
\vspace{0.2 cm}
\begin{tabular}{ l |  c | c | c | c | c }
\toprule
 Model &  Latency [$\mu$s]  &  $R=(6,5)$ [$\mu$s] &  $R=(12,10)$ [$\mu$s] &  $R=(30,20)$ [$\mu$s] &  $R=(60,60~[40])$ [$\mu$s]  \\
 \midrule
 GRU &  1.7--1.7  &  2.4--6.5  &  3.2--7.3  &  5.0--9.1  &  8.0--12.1  \\ 
 LSTM &  1.7--1.7  &  2.7--6.8  &  3.5--7.6  &  5.3--9.4  &  8.3--12.4  \\ 
 \bottomrule
\end{tabular}
\end{table}

\begin{table}[h!]
\centering
\caption{ Minimum and maximum latencies for the jet flavor tagging model.}
\label{tab:flavortag}
\vspace{0.2 cm}
\begin{tabular}{ l | c | c | c | c }
\toprule
 Model &  $R=(48,40)$ [$\mu$s]  &  $R=(90,60)$ [$\mu$s] &  $R=(120,120)$ [$\mu$s]  &  $R=(240,240)$ [$\mu$s] \\
 \midrule
 GRU &  6.7--24.8  &  9.8--27.9 &  11.5--29.6  &  20.5--38.6 \\ 
 LSTM &  6.9--25.0 &  10.1--28.2  &  11.7--29.8 &  20.7--38.8 \\
 \bottomrule
\end{tabular}
\end{table}

\begin{table}[h!]
\centering
\caption{ Minimum and maximum latencies for the QuickDraw model.}
\label{tab:quickdraw}
\vspace{0.2 cm}
\begin{tabular}{ l | c | c | c | c }
\toprule
 Model &  $R=(48,32)$ [$\mu$s] &  $R=(96,64)$ [$\mu$s] &  $R=(192,128)$ [$\mu$s] &  $R=(384,384~[256])$ [$\mu$s] \\
 \midrule
 GRU &  35.4--164.0  &  59.4-–188.0 &  107.0--235.0  &  203.0--331.0 \\ 
 LSTM &  35.9--164.0  &  59.9--188.0  &  107.0--236.0  &  203.0--332.0  \\ 
 \bottomrule
\end{tabular}
\end{table}

\subsection{Static and Non-static Comparison}

\begin{figure}[h!]
\centering

    \begin{subfigure}[b]{0.33\textwidth}
    \centering
    \includegraphics[width=\textwidth]{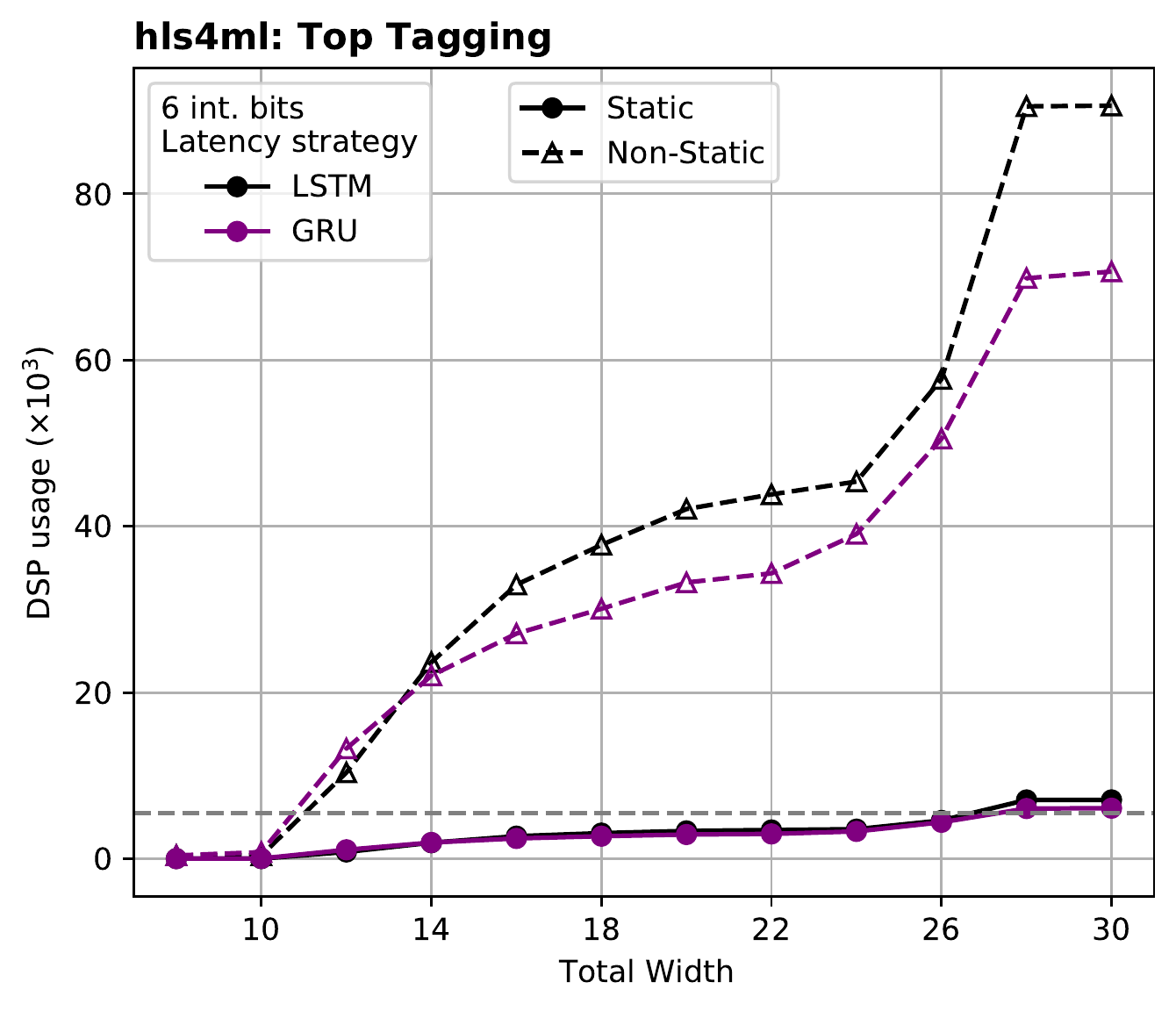}
    \caption{Top quark tagging: DSP}
    \end{subfigure}
    \begin{subfigure}[b]{0.33\textwidth}
    \centering
    \includegraphics[width=\textwidth]{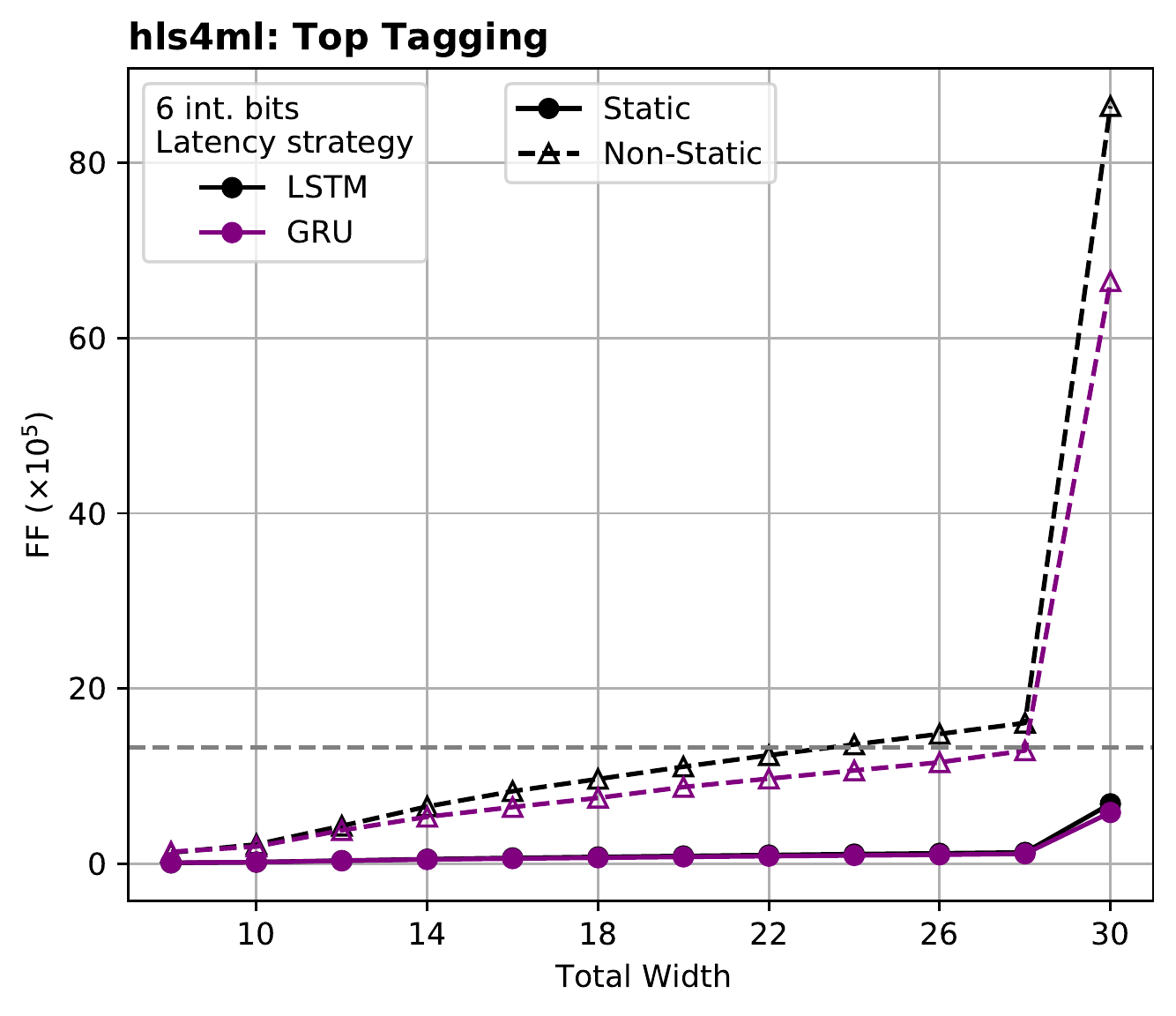}
    \caption{Top quark tagging: FF}
    \end{subfigure}
    \begin{subfigure}[b]{0.33\textwidth}
    \centering
    \includegraphics[width=\textwidth]{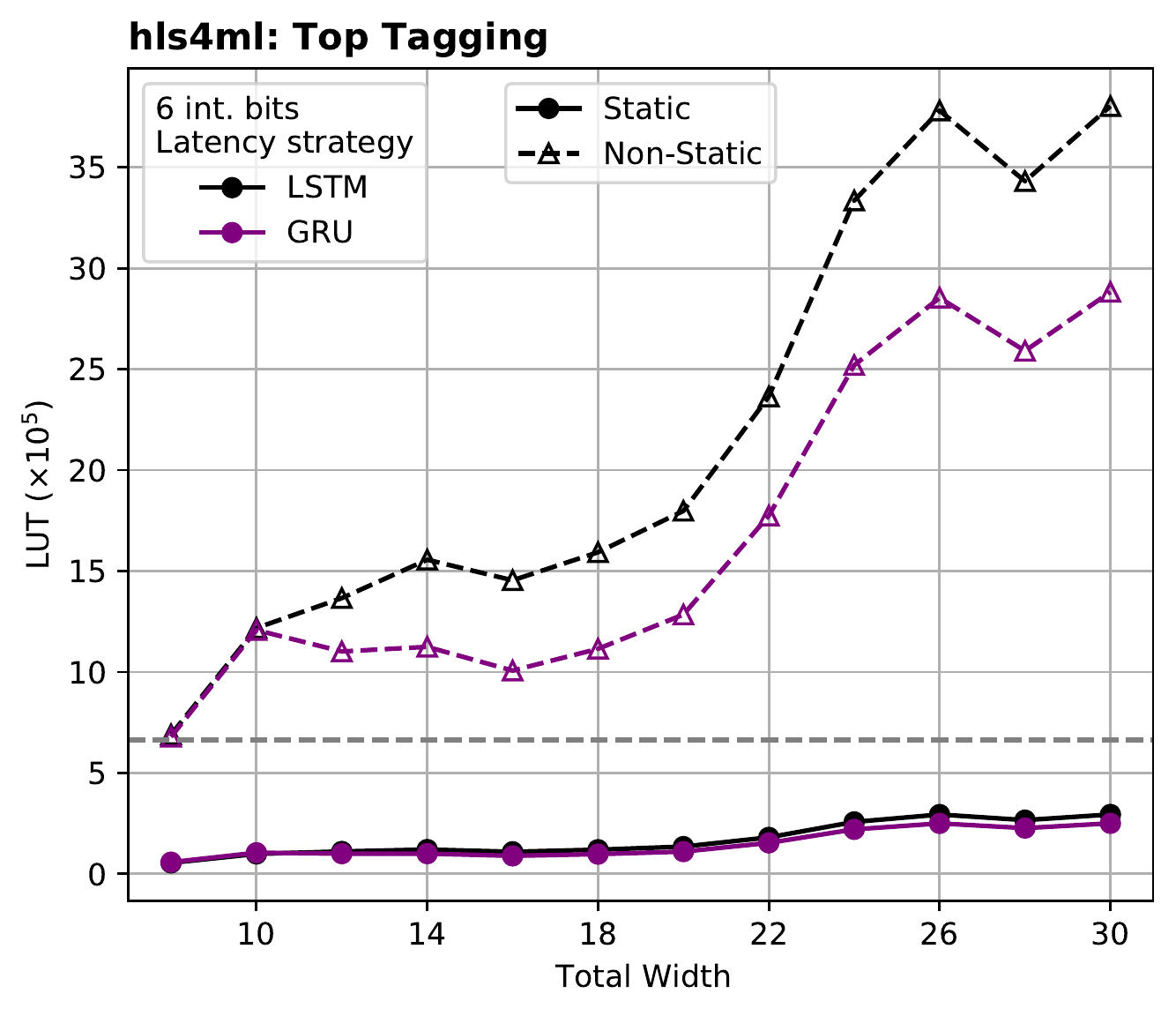}
    \caption{Top quark tagging: LUT}
    \end{subfigure}
    
\caption{Resource usage of DSPs (a), FFs (b), and LUTs (c) for the top quark tagging model in static and non-static mode with solid and dashed lines, respectively.
The resources available in the target FPGA are shown in the black dashed horizontal line.}
\label{fig:util_staticnonstatic}
\end{figure}

In order to study the impacts of the static and non-static modes discussed in Sec~\ref{sec:impl} we limit our consideration to the top quark tagging models.
As shown in Fig.~\ref{fig:util_staticnonstatic}, resource usage for non-static mode increases dramatically compared to static mode.
For even moderate-sized models, non-static mode requires too many resources to be feasible.
In the case of the top quark tagging model we see that non-static mode is able to fit within the available resources of the chip only for very small bitwidths.
However, Table~\ref{tab:staticnonstatic} confirms that although non-static mode offers similar overall latency to static mode, the II in non-static mode is reduced from 315 (314) to 1 for the GRU (LSTM) models.
This results in a increased throughput for non-static mode by a factor of more than 300.
The increased throughput of non-static mode would be vital for L1 trigger applications that run inferences at rates of up to 40\,MHz.
While this particular top quark tagging model suffers in performance using a total bitwidth of 10 (6 integer and 4 fractional bits in this case), there are multiple options, such as per-layer quantization or quantization-aware training, that were not considered for this study but could potentially allow a performant version of this model to be synthesized in non-static mode.

\begin{table}[h!]
\centering
\caption{ Minimum and maximum latencies and initiation intervals for the top quark tagging model in both static and non-static mode.}
\label{tab:staticnonstatic}
\vspace{0.2 cm}
\begin{tabular}{ l | c | c | c | c }
\toprule
 Model &  Static latency [$\mu$s] &  Non-static latency [$\mu$s]  &  Static II  &  Non-static II   \\
 \midrule
GRU  &  1.7--1.7  &  1.6--1.6  &  315 &  1 \\ 
LSTM &  1.6--1.6  &  1.5--1.5  &  314 &  1 \\ 
 \bottomrule
\end{tabular}
\end{table}

%% file: text/09_summary.tex
\section{Summary and Outlook}

RNNs have shown substantial success for many tasks in particle physics.
They are particularly well-suited to those problems involving sequences of particle or detector signals, outperforming densely connected deep neural networks (DNNs) \cite{Egan:2017ojy} and convolution neural networks (CNNs) \cite{Fraser:2018ieu} on certain jet classification tasks. 
In spite of this success, RNNs have not seen the widespread adoption in ultra-low latency environments in physics when compared to DNNs and CNNs. 
This difference is owed in part to tools such as hls4ml that simplify the adaptation of the latter models from Keras to HLS.
The support for GRUs and LSTMs in hls4ml that we present in this work represents the removal of a major barrier to the use of RNNs in ultra-low latency environments.
This has ramifications not only for high energy physics but also other research areas where RNNs have become popular.
While we have focused on the usage of hls4ml with FPGAs, it is important to note that hls4ml can also be used to create ASIC designs~\cite{DiGuglielmo:2021ide}, and thus this work also allows for the possibility of RNN usage on ASICs as well.

The implementation we present in hls4ml in this work maintains the main tuning features of hls4ml, namely the reuse factor and per-layer bit precision.
This is necessary to allow the customization of the synthesized design to meet the needs of a given task.
The benchmark models chosen cover a range of sizes, latencies, and problems, and showcase the quality of the hls4ml support for a variety of realistic scenarios.
We also add an RNN-specific tuning parameter to hls4ml called the RNN mode, with static and non-static settings capable of further adjusting the behavior of the synthesized design.
While we show that this work is capable of producing results with high accuracy, there are multiple possibilities for future development.
In particular, we observe that even small RNN models can require a substantial amount of resources to implement.
While the post-training quantization scheme we have used here is able to minimize resources to a certain extent, other methods, such as quantization-aware training, have shown that even more resource reduction can be possible with little to no cost to performance.
This is perhaps even more true for RNNs than dense neural networks due to the repeated use of the recurrent layer weights.
Other techniques such as masking are also a possible method for reducing both resource usage and dependence on small weight values (high bit precision).

The recurrent or repeating nature of many modern algorithms, such as RNNs, transformers and graph neural networks, make them very difficult to be run, particularly at low latency, on FPGAs. 
In this work, we present the successful deployment of RNNs in models with number of trainable parameters ranging from $\mathcal{O}$(1\,k) to $\mathcal{O}$(100\,k) achieving latencies of $\mathcal{O}$(1\,$\mu$s) to $\mathcal{O}$(100\,$\mu$s). 
This represents an important step in enabling support in hls4ml for more complex architectures with recursive computations.

%% file: text/10_acknowledgment.tex
We acknowledge the Fast Machine Learning collective as an open community of multi-domain experts and collaborators.
This community was important for the development of this project. We thank Javier Duarte, Maurizio Pierini, Zhiqiang Que and Nhan Tran for their valuable comments and suggestions.
M. Kagan and R. Teixeira de Lima are supported by the US Department of Energy (DOE) under grant DE-AC02-76SF00515.
P. Harris and D. Rankin acknowledge DOE grant DE-SC0021943 and National Science Foundation (NSF) grants No. 1934700 and 1931469.
S.-C. Hsu and E. Khoda are supported by NSF grants No.2117997 and 1934360.

%% file: text/11_code_availability.tex
The hls4ml library is available at \href{https://github.com/fastmachinelearning/hls4ml}{https://github.com/fastmachinelearning/hls4ml} and RNN support is available as of commit \href{https://github.com/fastmachinelearning/hls4ml/commit/59ed8249f4bbdb4b23ff0c6f0bfc976b44d3ac7e}{\texttt{59ed8249f4bbdb4b23ff0c6f0bfc976b44d3ac7e}}. For examples on how to use hls4ml, the notebooks in \href{https://github.com/fastmachinelearning/hls4ml-tutorial}{https://github.com/fastmachinelearning/hls4ml-tutorial} serve as a general introduction.